\documentclass[conference]{IEEEtran}
\IEEEoverridecommandlockouts

\IEEEoverridecommandlockouts
\usepackage{booktabs} % For formal tables
\usepackage{subfiles}
\usepackage{amsmath,amssymb,amsfonts}
\usepackage{algorithm}
\usepackage{algpseudocode}
\usepackage{graphicx}
\usepackage{xcolor}
\usepackage{color}
\usepackage[switch]{lineno}
\usepackage[hidelinks]{hyperref}
\usepackage{mathtools}
\usepackage{amsthm}
\usepackage{float}
\usepackage{caption}
\usepackage{subcaption}
\def\BibTeX{{\rm B\kern-.05em{\sc i\kern-.025em b}\kern-.08em
    T\kern-.1667em\lower.7ex\hbox{E}\kern-.125emX}}

\newcommand{\etal}{\textit{et~al}.}

\begin{document}

\title{Optimizing Real-Time Performances for Timed-Loop Racing under F1TENTH \\
\thanks{We thank Zhaojian Li from Michigan State University and Marko Bertogna from University of Modena for fruitful discussions on the topic. We also thank Nvidia for providing the Jetson Xavier AGX used in this paper.}
}
% NVIDIA or Nvidia? They mention Nvidia everywhere, only the brand logo is all-cap I think

\author{\IEEEauthorblockN{Nitish A Gupta}
\IEEEauthorblockA{North Carolina State University\\
Department of Computer Science\\
%Raleigh, NC\\
nagupta@ncsu.edu}
\and
\IEEEauthorblockN{Kurt Wilson}
\IEEEauthorblockA{University of Central Florida\\
Department of Computer Science\\
%Orlando, FL\\
kurt.wilson@knights.ucf.edu}
\and
\IEEEauthorblockN{Zhishan Guo}
\IEEEauthorblockA{North Carolina State University\\
Department of Computer Science\\
%Raleigh, NC\\
zguo32@ncsu.edu}}

\maketitle
\thispagestyle{plain}
\pagestyle{plain}

\begin{abstract}
Motion planning and control in autonomous car racing are one of the most challenging and safety-critical tasks due to high speed and dynamism. The lower-level control nodes are expected to be highly optimized due to resource constraints of onboard embedded processing units, although there are strict latency requirements. Some of these guarantees can be provided at the application level, such as using ROS2's Real-Time executors. However, the performance can be far from satisfactory as many modern control algorithms (such as Model Predictive Control) rely on solving complicated online optimization problems at each iteration. In this paper, we present a simple yet effective multi-threading technique to optimize the throughput of online-control algorithms for resource-constrained autonomous racing platforms. We achieve this by maintaining a systematic pool of worker threads solving the optimization problem in parallel which can improve the system performance by reducing latency between control input commands. We further demonstrate the effectiveness of our method using the Model Predictive Contouring Control (MPCC) algorithm running on Nvidia's Xavier AGX platform.
\end{abstract}

% \begin{IEEEkeywords}
% MPCC, ROS2, F1Tenth, Real-Time, Opimization, Embedded, Autonomous Racing
% \end{IEEEkeywords}
\section{Introduction}\label{sec:introduction}

High-speed autonomous racing is a challenging and important application within autonomous driving, wherein safety is a crucial consideration. In recent years, this application has gained increased attention from the real-time systems community. Autonomous navigation at high-speed demands tighter bounds on response times of the localization and control algorithms. Several competitions such as F1TENTH \cite{o2020f1tenth} and Indy Autonomous Challenge (IAC) \cite{iac_paper} are held once to twice every year to validate and improve the functional limits of the existing software stack. The F1TENTH competition focuses on racing 1/10th-scale vehicles in a dynamic head-to-head racing environment. At this scale, the problem of autonomous navigation narrows down to designing an efficient and real-time safe trajectory following an algorithm in a dynamic environment. To this end, different control algorithms \cite{betz} are employed by the community such as Reactive, Model Predictive Control (MPC) \cite{mpcracing}, raceline-optimization \cite{tunercar} or Learning-based algorithms \cite{turismo}. In this paper, we focus our attention on an online optimization-based MPC algorithm, particularly, Model Predictive Contouring Control (MPCC). MPCC algorithm provides high-performance autonomous navigation accounting for non-linearities in dynamics efficiently at high speed. This is mainly because the MPCC solver accounts for model deviations at each step and applies relevant corrections dynamically. Therefore, the closed-loop online optimization as in MPCC particularly tends to be more reliable than other algorithms mentioned earlier. In the context of autonomous racing, this algorithm was originally demonstrated in \cite{liniger2015optimization} on 1:43 scale RC cars. 

\begin{figure}
\centering
\includegraphics[trim=0 1.2in 0 0, clip, width=0.7\linewidth]{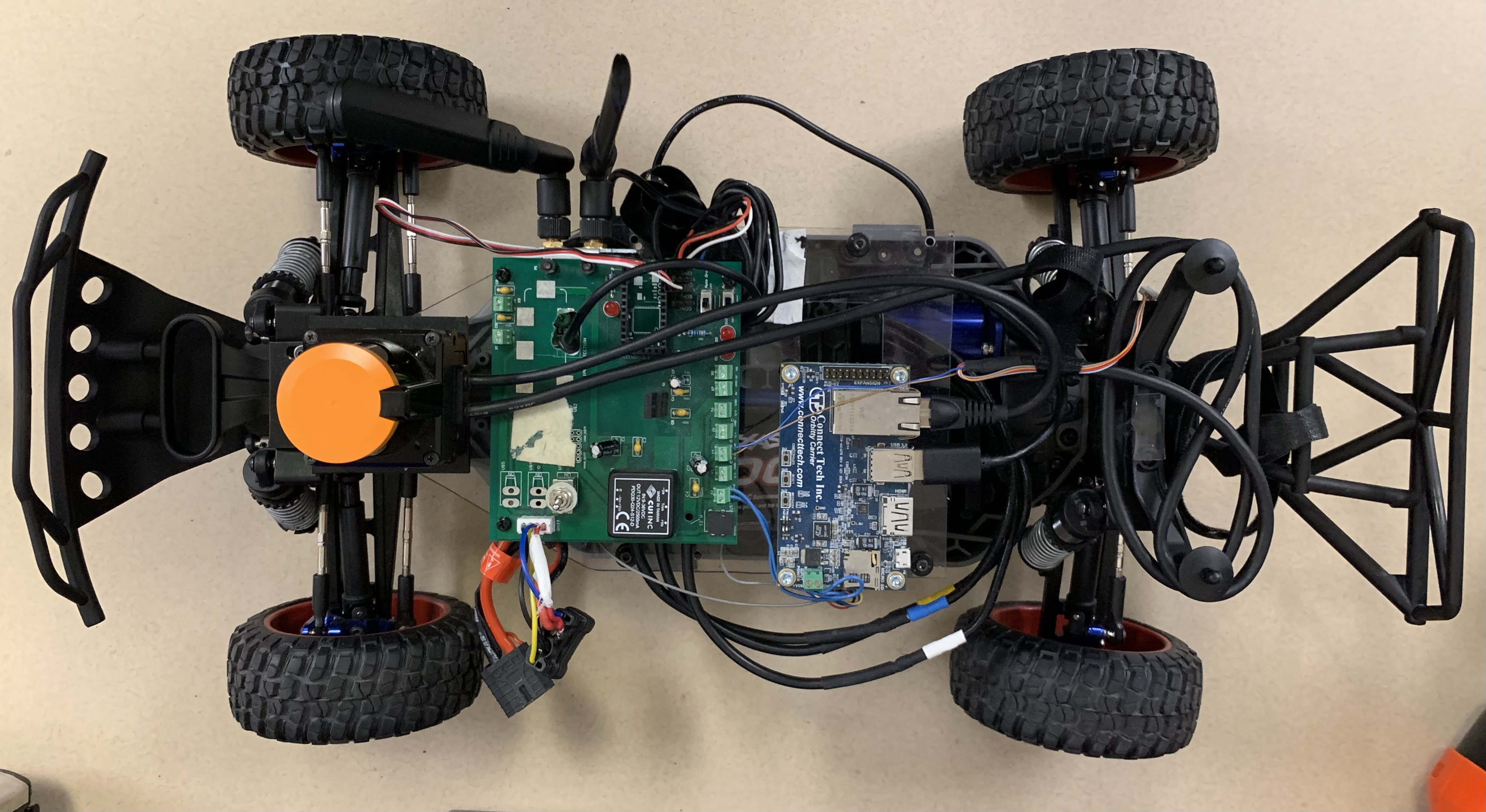}
\caption{F1TENTH Platform}
\label{fig:f110}
\vspace{-4mm}
\end{figure}

As compared to Autonomous Vehicle (AV) applications for urban driving, autonomous vehicle racing platform pose a different set of challenges such as frequent edge cases, resource constraints, and critical reaction/response time. Thus it becomes highly difficult to deploy algorithms that rely on online optimization under these constraints. However, control algorithms such as MPCC rely on solving a constrained convex optimization problem over a finite horizon to provide a dynamic and efficient navigation strategy.

To apply such a technique for high-speed autonomous navigation on embedded platforms such as Nvidia Xavier NX/AGX, the latency between actuation command signals needs to be optimized from both the application and system perspective. We discuss this problem in the next sections and present a simple yet elegant approach for reducing this latency of online optimization control algorithms for timed-loop autonomous racing platforms. In particular, we propose a multi-threading based algorithm on top of the ROS2 implementation of MPCC, such that the online optimization can be performed by multiple threads in a synchronized fashion. This enables us to improve the publish latency of the control algorithm while optimizing the CPU utilization in a resource-constrained environment.

% \textcolor{blue}{shall we list main contributions for industry paper?}
% The main contributions of this paper are, 

% \begin{enumerate}
%     \item Optimizing the MPCC control loop timing through synchronized multi-threaded controller node
%     \item Evaluating performance of MPCC control node in ROS2 Real-Time executors
% \end{enumerate}

% The remainder of the paper is organized as follows. Section~\ref{sec:background} presents a brief background of F1tenth platform and software stack, Model Predictive Contouring Control algorithm, and ROS2 architecture. Section~\ref{sec:problem} describes the problem statement, followed by Section~\ref{sec:implementation}, wherein we provide a detailed implementation and proposed algorithm. In Section~\ref{sec:experiments}, we evaluate and demonstrate our proposed optimization approach. Finally, in Section~\ref{sec:conclusion}, we present our conclusions and future work.

% background moved to background.tex
\section{Background}\label{sec:background}

This section briefly explains ROS2 architecture and then provides the details of Model Predictive Contouring Control, as well as the F1TENTH hardware and software stack.

\subsection{Architecture of ROS2}
ROS2 is a popular robotics framework that leverages the publisher-subscriber paradigm to provide real-time inter-process communication via the data distribution service layer in a distributed computing environment. The simplest form of computing unit in ROS2 is called \textit{nodes} which may in turn contain several callbacks. Each callback in itself is an executable function that could be either time-triggered or data-triggered. The message transfer in ROS is handled through a \textit{topic}. A node can publish and/or subscribe to one or several topics. A subscriber callback of a node is triggered whenever the topic it subscribes to receives a message from a publisher. A node publishes a topic using a timer which is used to set the rate at which the timer callback will be triggered and thus the message will be published. Beneath the abstraction, all callbacks within a single node are scheduled for execution by an executor. The default ROS2 executor behavior has been thoroughly studied \cite{casini2019response,tang2020response} and several customized schedulers also have been proposed such as dynamic-priority~\cite{arafat2022response} and fixed-priority~\cite{choi2021picas} schedulers for executors.

\subsection{Model Predictive Contouring Control}\label{subsec:algoMPCC}
Model Predictive Control is a popular optimal control algorithm used for process control and trajectory following in linear and non-linear systems. MPC achieves the prediction of multiple control inputs over a finite receding horizon by solving a constrained convex optimization problem. For optimal performance, the cost function formulation must capture the process goals and the constraints from the model and environment. In the case of autonomous racing, the cost function can be represented by the progress along the race line while maintaining a safe distance from the track borders and obstacles. At each time step, the first predicted input is applied and the optimization is repeated over the new horizon.

In this paper, we consider a reconfigured and optimized implementation of Model Predictive Contouring Control by Liniger~\etal~\cite{liniger2015optimization}. However, instead of a 1/43rd scale, we model it for a 1/10th scale by updating the vehicle dynamics for F1TENTH using the methodology described in \cite{orca_f110}. The state representation is based on the vehicle bicycle model and control input consists of duty cycle $d$ and steering angle $\delta$. 

\vspace{-2mm}

\begin{equation}\label{def:xdot}\dot{X} = v_x \cos{\varphi} - v_y\sin{\varphi}\end{equation}
\begin{equation}\label{def:ydot}\dot{Y} = v_x \sin{\varphi} + v_y\cos{\varphi}\end{equation}
\begin{equation}\label{def:phidot}\dot{\varphi} = \omega\end{equation}
\begin{equation}\label{def:vxdot}\dot{v_x} = (F_{r,x} - F_{f,y}\sin{\delta} + mv_y\omega)/m\end{equation}
\begin{equation}\label{def:vydot}\dot{v_y} = (F_{r,y} + F_{f,y}\cos{\delta} - mv_x\omega)/m\end{equation}
\begin{equation}\label{def:wdot}\dot{\omega} = (F_{f,y}l_f\cos{\delta} - F_{r,y}l_r)/I_z\end{equation}

In the above differential equations, $X$, $Y$, and $\varphi$ is vehicle's position and heading angle relative to the inertial frame. Similarly, the $v_x$, $v_y$, and $\omega$ represent linear and angular velocities. The $F_f$ and $F_r$ represent tire forces along the lateral and longitudinal direction which are modeled using a simplified Pacejka Tire Model. Finally, $m$, $l_r$, and $l_f$ is vehicle mass and front/rear distance from vehicle's $COG$, respectively. The dynamics modeling part is beyond the scope of this paper and can be found in \cite{liniger2015optimization, orca_f110}. Considering the goal of maximizing the progress measured along the parameterized centerline and satisfying several equality and inequality constraints, a convex optimization problem is formulated. The online optimization is solved using a high-performance Quadratic Programming (QP) solver over a finite horizon $N$, such that all constraints are satisfied. Only the first input from the entire horizon solution is applied as input commands to the vehicle. This process is then repeated in the next cycle over the entire horizon. This ensures the stability of the control algorithm and keeps up with slight inconsistencies in vehicle dynamics modeling. The horizon length parameter, \textit{N}, controls the tracking accuracy as it provides farsightedness to the vehicle and allows for better planning. 

\subsection{F1TENTH Platform}\label{subsec:f1tenth}

\begin{figure}[h]
\centering
\includegraphics[width=\linewidth]{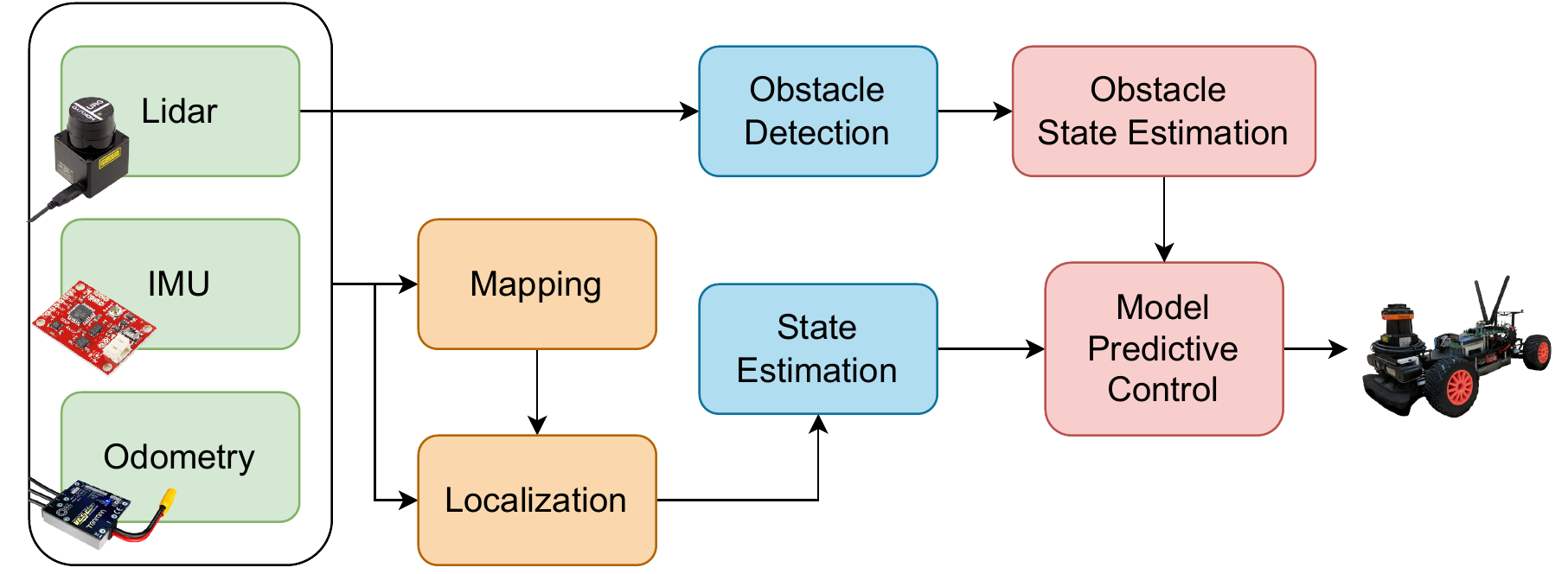}
\caption{F1TENTH Software Stack}
\label{fig:f110_stack}
\vspace{-1mm}
\end{figure}

The F1TENTH platform is built upon a 1/10th scale chassis which hosts a differential 4-wheel drive and Ackermann steering mechanism. The steering and throttle commands are sent to the motors through the VESC 6 MkV controller which also allows for motor feedback. An embedded platform such as Nvidia Jetson AGX Xavier is used as an onboard processing unit. The sensor stack consists of a Hokuyo UST-10LX Lidar, a 9-DoF IMU, and an optional ZED Depth camera. The Lidar has a detection range of 10 meters, a scan range of 270$^{\circ}$ at 0.25$^{\circ}$ angular resolution, and a scan speed of 40 Hz. The Lidar along with IMU sensor and odometry data enables mapping and localization of the ego vehicle using algorithms such as Google's cartographer~\cite{hess2016real} or slam-toolbox \cite{slam}. Fig.~\ref{fig:f110_stack} shows the essential modules and data flow in a typical F1TENTH platform. The communication between the sensors, embedded board, and speed controllers is facilitated using ROS2.

The sensor stack and localization module enable state estimation of the ego and opponent vehicle, which is then communicated to the MPCC algorithm for calculating a control input command. The MPCC controller solves and publishes the control input command consisting of $(d, \delta)$, to the lower level VESC and steering controller nodes for actuation of the vehicle. Upon actuation, the state estimation repeats and transitions the system to the next state. However, the applied target input command is not achieved instantly due to the physical constraints of the vehicle. Thus the updated state is not necessarily a linear interpolation of the applied control input, but instead represents an integration of model dynamics over small finite intervals. As we will see in the next section, we try to capture this behavior using the iterative \textit{Runga-Kutta} (RK4) numerical method. 

\section{Problem Statement}\label{sec:problem}
\begin{figure} [h]
\centering
\includegraphics[width=0.75\linewidth]{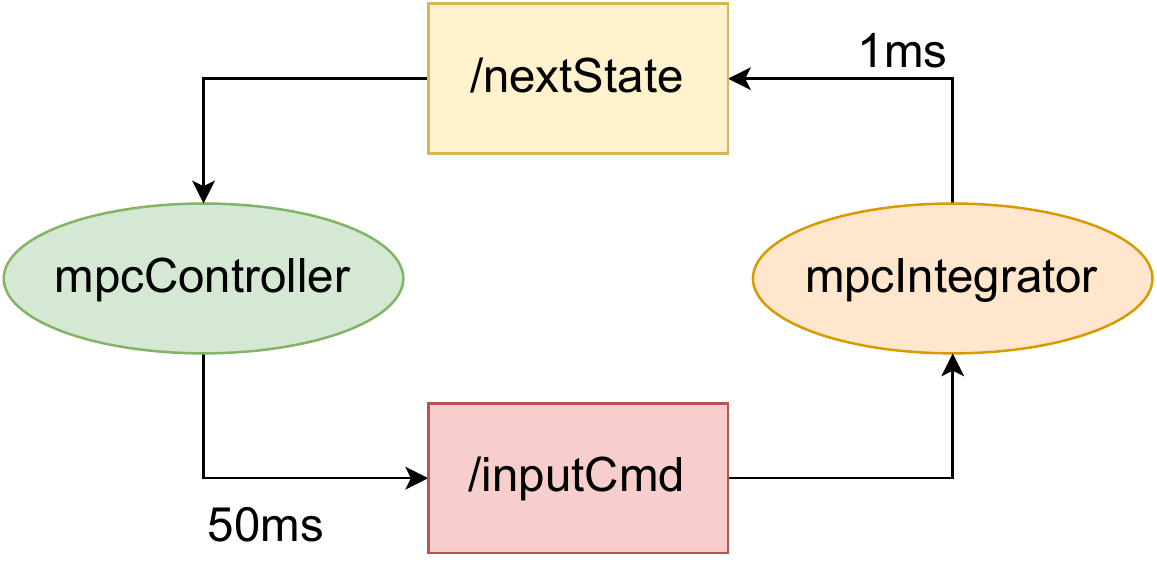}
\caption{Simplified MPCC Control Loop}
\vspace{-2mm}
\label{fig:mpc_ros2_1}
\end{figure}

As the main focus of this paper is optimizing the control performance, we consider a simplified representation of the framework explained in Section \ref{sec:background} as two main nodes implemented in ROS2. As shown in Fig. \ref{fig:mpc_ros2_1}, the first node is the \textit{mpcController} node that initializes and solves an online convex optimization problem. The \textit{mpcController} node subscribes to vehicle state on the \textit{nextState} topic and publishes the vehicle control input commands to \textit{inputCmd} topic. The second node on the control loop is \textit{mpcIntegrator} node which represents the model state integrator. The \textit{mpcIntegrator} node subscribes to control inputs commands on \textit{mpcInput} topic and transitions the vehicle to the next state which is then published to \textit{nextState} topic. On a physical F1TENTH platform, the role of the state integrator is distributed between actuating the vehicle and then estimating the next state through the localization module and other sensors such as IMU and Odometry. As the combined sensor stack on the F1TENTH platform publishes data at a high frequency ranging anywhere from 40 Hz to 100 Hz, the state updates are processed faster than the control optimization. This behavior is captured in our implementation by executing a state simulator thread within \textit{mpcIntegrator} node at 1000 Hz or \textit{1 ms} interval. The state integrator updates the states of the vehicle using the most recently available control input command, thus it is important that the \textit{mpcController} node publishes new control inputs at a high frequency to maintain data freshness. 

% removed
% It is worth noting that the state integrator updates the states of the vehicle using the most recently available control input command. Thus it becomes imperative for the \textit{mpcController} node to publish new control inputs at high frequency to maintain data freshness. 

% \begin{figure}
% \centering
% \includegraphics[width=0.9\linewidth]{figures/MpcRos2_mod.pdf}
% \caption{Multi-threaded Control Loop \textcolor{blue}{[probably will remove/update this one]}}
% \label{mpc_ros2_2}
% \end{figure}
\section{Implementation}\label{sec:implementation}

The \textit{mpcController} node operates on the current vehicle state to solve a convex optimization problem over a finite horizon $N$. The controller then publishes the first control input from the solved horizon. As the autonomous racing control problem is highly dynamic, the time interval between each new command message is crucial to optimize to keep the vehicle as reactive as possible. From the application perspective, the optimization solver is only dependent on the vehicle's current state. On the other hand, the \textit{mpcIntegrator} node updates the system state at a much higher frequency and publishes on the \textit{nextState} topic. This gives us an opportunity to parallelize the controller node using multi-threading, wherein each worker thread is solving optimization based on the most recently available vehicle state. However, the control input commands that are based on two very similar states do not provide much benefit to overall system throughput. Similarly, a new command message based on an old vehicle state needs to be discarded when a command message based on a newer state has already been published. Such cases need to be handled in a systematic and synchronized manner to ensure reliability and data freshness. 

\begin{algorithm}
  \caption{Multi-threaded Controller Node}
  \label{alg1}
  \begin{algorithmic}[1]
    \State $last\_queued\_timestamp \gets 0$
    \State $last\_output\_time \gets 0$
    \State declare $atomic\_msg$
    \State declare $result\_msg$
    \State declare $worker\_thread\_cv$
    \State declare $publish\_thread\_cv$

    \Function{Process\_State\_Message}{State, Time}
        \State /* if enough time has passed */
        \If{$State.timestamp - last\_queued\_timestamp > MIN\_GAP$} 
        % $atomic_\msg.exchange(State)$\\
        \State atomic\_msg.exchange(State)
        \State /* notify the worker threads */
        \State worker\_thread\_cv.notify\_one()
        \EndIf
    \EndFunction
    
    \Function{workerthread}{}
    \While {should\_run}
        \State worker\_thread\_cv.wait()
        \State State = atomic\_msg.exchange(nullptr)
        \State /* update the shared timer */
        \State last\_queued\_timestamp = State.timestamp
        \State result = mpc(State)
        
        \State /* make sure another worker did not complete with a newer state */
        \If{State.timestamp $<$ last\_output\_time}
            \State break
        \EndIf
        \State last\_output\_time = State.timestamp;
        \State result\_msg = result;
        \State publish\_thread\_cv.notify\_one()
    \EndWhile
    \EndFunction
    
    \Function{publishthread}{}
    \While{true}
        \State publish\_thread\_cv.wait()
        \State publish(result\_msg)
    \EndWhile
    \EndFunction
  \end{algorithmic}
\end{algorithm}

We create a thread pool to run the \textit{mpcController} on different \textit{nextState} messages in parallel. The state subscriber callback of \textit{mpcController} node checks the timestamp of each incoming message and compares it to the timestamp of the last state message accepted by the thread pool. If the timestamp of an incoming message is too close to the most recently accepted state message by any thread, the received message is discarded. We call this interval $MIN\_GAP$, and the above check occurs on line-9 of Algorithm \ref{alg1}. This condition prevents the thread pool from being saturated with state messages from consecutive timestamps and ensures a well-spaced interval between published control input messages. The unassigned worker threads are required to wait until a state message arrives with a qualified timestamp. The next received state message is then passed to one of the available worker threads in the pool. The assigned thread updates the shared timestamp and starts working on the MPCC optimization call. Once the thread completes the call, it checks whether another worker has already published a result based on newer state data. If not, it wakes the publisher thread with the new control input message to publish on \textit{mpcInput} topic. A global state variable $last\_output\_time$, maintains the timestamp of the last used state data to ensure that no control message is sent based on older state messages. This check occurs on line-23 of the algorithm. This extra check must occur since a thread working on state data could be delayed due to operating system preemptions. During this time another thread could have finished with newer state data and thus invalidating the control message obtained from the delayed worker thread.

To ensure that the worker threads do not interfere with each other, we pin each thread to a CPU core and assign each thread a high priority.

\section{Preliminary Evaluation}\label{sec:experiments}
\begin{figure}
\centering
    \begin{subfigure}[b]{0.45\textwidth}
    \centering
        \includegraphics[width=\linewidth]{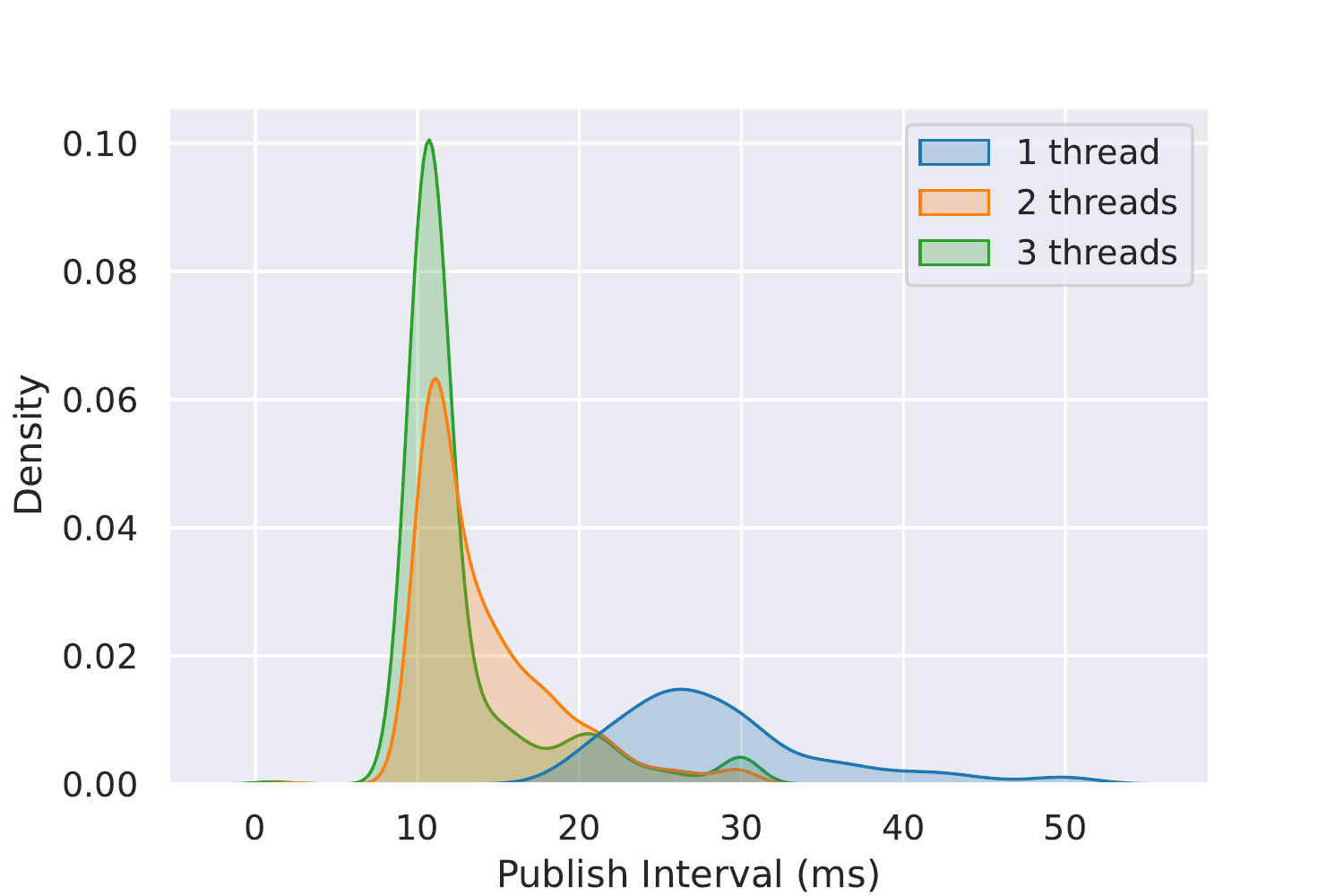}
        \caption{Distribution of the MPC control output interval as more threads are added to the worker pool. Adding threads both reduces the mean and variance of the interval}
        \label{fig:plota}
    \end{subfigure}
    \begin{subfigure}[b]{0.45\textwidth}
    \centering
        \includegraphics[width=\linewidth]{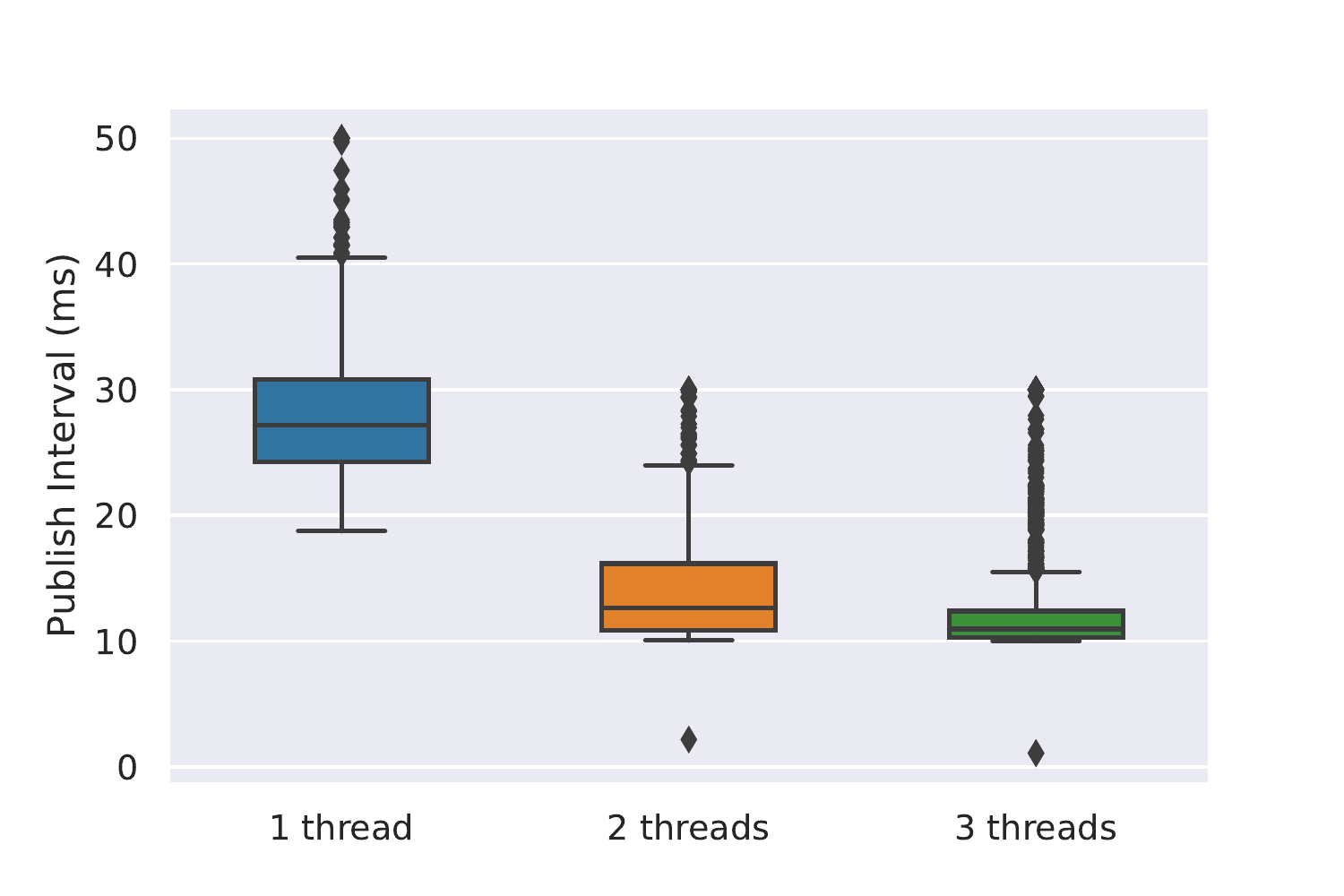}
        \caption{Ranges of the MPC control output intervals. Using three threads allows the control interval to stay much closer to the enforced minimum of 10ms. Note that there are cases where the interval was less than 10ms; this can happen when two threads started 10ms apart but completed around the same time and in order}
        \label{fig:plotb}
    \end{subfigure}
    \caption{Results from testing the threaded MPC algorithm on a Nvidia Jetson AGX Xavier}
\label{fig:plots}
\end{figure}

In this section, we present an initial evaluation to show the efficacy of our proposed optimization method. We model the C++ implementation of MPCC algorithm using the latest version of ROS2 Humble, based on the representation described in Section \ref{sec:problem}. We performed our experiments on Nvidia Jetson AGX Xavier running Ubuntu 20.04 with Linux Real-Time kernel. We set $MIN\_GAP$ parameter set to 10 ms, that is no worker thread can be started within 10 ms of another worker. Therefore, the target publish interval is 10 ms. The evaluation results are shown in Fig. \ref{fig:plota} and Fig. \ref{fig:plotb}.

% things to highlight:
% significant decrease in average and max publish interval
% with a enforced minimum interval of 10ms, three threads achieves within 6 ms of the target spacing, while one thread may wait 40ms between results, reducing the freshness of control data
% decrease in publish interval variance

% am I just stating the obvious here?
The experiments show that adding additional workers significantly reduces the maximum and average control publish interval length. This enables vehicle actuation on more recent and frequent input command signal. Adding a second thread brings the absolute maximum rate from 50 ms down to 30 ms. The three-thread scenario keeps the maximum publishing rate much closer to the target rate than with just one thread, demonstrating the benefits of synchronized parallel processing. While the maximum observed publishing rates are close for two and three threads, the variance of the publishing interval is significantly decreased when using three threads. % increasing predictability / reducing the chance of outliers?
% the average and 95\% publish rate is reduced when using three threads.

% is there a better way to express that 95% percentile thing? is it worth mentioning? maybe not?

% \begin{figure}
% \centering
% \caption{}
% \label{exp2}
% \end{figure}
\section{Conclusion}\label{sec:conclusion}
In this paper, we presented a challenge that is commonly faced while developing optimization algorithms for autonomous vehicle racing. We then presented a multi-threading-based solution to optimize the control throughput performance of online optimization algorithms on resource-constrained platforms such as F1TENTH. In the future, we plan to investigate other ways of optimizing the performance by utilizing the shared GPU memory on embedded platforms for performing expensive solver matrix calculations. Finally, we plan to demonstrate our proposed algorithm during an upcoming F1TENTH competition.

\bibliographystyle{ieeetr}
\bibliography{references}

\begin{thebibliography}{10}

\bibitem{o2020f1tenth}
M.~O'Kelly {\em et~al.}, ``{F1TENTH: An Open-source Evaluation Environment for
  Continuous Control and Reinforcement Learning},'' {\em Proceedings of Machine
  Learning Research}, vol.~123, 2020.

\bibitem{iac_paper}
A.~Wischnewski {\em et~al.}, ``Indy autonomous challenge - autonomous race cars
  at the handling limits,'' in {\em 12th International Munich Chassis Symposium
  2021} (P.~Pfeffer, ed.), (Berlin, Heidelberg), pp.~163--182, Springer Berlin
  Heidelberg, 2022.

\bibitem{betz}
J.~Betz, H.~Zheng, A.~Liniger, U.~Rosolia, P.~Karle, M.~Behl, V.~Krovi, and
  R.~Mangharam, ``Autonomous vehicles on the edge: A survey on autonomous
  vehicle racing,'' {\em IEEE Open Journal of Intelligent Transportation
  Systems}, vol.~3, pp.~458--488, 2022.

\bibitem{mpcracing}
E.~Thil{\'e}n, ``Robust model predictive control for autonomous driving,''
  2017.

\bibitem{tunercar}
M.~O’Kelly, H.~Zheng, A.~Jain, J.~Auckley, K.~Luong, and R.~Mangharam,
  ``Tunercar: A superoptimization toolchain for autonomous racing,'' in {\em
  2020 IEEE International Conference on Robotics and Automation (ICRA)},
  pp.~5356--5362, IEEE, 2020.

\bibitem{turismo}
P.~R. Wurman, S.~Barrett, K.~Kawamoto, J.~MacGlashan, K.~Subramanian, T.~J.
  Walsh, R.~Capobianco, A.~Devlic, F.~Eckert, F.~Fuchs, {\em et~al.},
  ``Outracing champion gran turismo drivers with deep reinforcement learning,''
  {\em Nature}, vol.~602, no.~7896, pp.~223--228, 2022.

\bibitem{liniger2015optimization}
A.~Liniger {\em et~al.}, ``Optimization-based autonomous racing of 1: 43 scale
  rc cars,'' {\em Optimal Control Applications and Methods}, vol.~36, no.~5,
  pp.~628--647, 2015.

\bibitem{casini2019response}
D.~Casini {\em et~al.}, ``Response-time analysis of ros 2 processing chains
  under reservation-based scheduling,'' in {\em ECRTS}, 2019.

\bibitem{tang2020response}
Y.~Tang {\em et~al.}, ``Response time analysis and priority assignment of
  processing chains on ros2 executors,'' in {\em RTSS}, IEEE, 2020.

\bibitem{arafat2022response}
A.~A. Arafat {\em et~al.}, ``Response time analysis for dynamic priority
  scheduling in ros2,'' in {\em DAC}, ACM/IEEE, 2022.

\bibitem{choi2021picas}
H.~Choi {\em et~al.}, ``Picas: New design of priority-driven chain-aware
  scheduling for ros2,'' in {\em RTAS}, IEEE, 2021.

\bibitem{orca_f110}
D.~Zahr{\'a}dka, ``Optimization-based control of the f1/10 autonomous racing
  car,'' 2020.

\bibitem{hess2016real}
W.~Hess {\em et~al.}, ``Real-time loop closure in 2d lidar slam,'' in {\em
  ICRA}, IEEE, 2016.

\bibitem{slam}
S.~Macenski and I.~Jambrecic, ``Slam toolbox: Slam for the dynamic world,''
  {\em Journal of Open Source Software}, vol.~6, no.~61, p.~2783, 2021.

\end{thebibliography}

\end{document}